\PassOptionsToPackage{unicode}{hyperref}
\PassOptionsToPackage{hyphens}{url}
\PassOptionsToPackage{dvipsnames,svgnames,x11names}{xcolor}
\documentclass[
  11pt,
]{article}
\usepackage{amsmath,amssymb}
\usepackage{iftex}
\ifPDFTeX
  \usepackage[T1]{fontenc}
  \usepackage[utf8]{inputenc}
  \usepackage{textcomp} % provide euro and other symbols
\else % if luatex or xetex
  \usepackage{unicode-math} % this also loads fontspec
  \defaultfontfeatures{Scale=MatchLowercase}
  \defaultfontfeatures[\rmfamily]{Ligatures=TeX,Scale=1}
\fi
\usepackage{lmodern}
\ifPDFTeX\else
\fi
\IfFileExists{upquote.sty}{\usepackage{upquote}}{}
\IfFileExists{microtype.sty}{% use microtype if available
  \usepackage[]{microtype}
  \UseMicrotypeSet[protrusion]{basicmath} % disable protrusion for tt fonts
}{}
\makeatletter
\@ifundefined{KOMAClassName}{% if non-KOMA class
  \IfFileExists{parskip.sty}{%
    \usepackage{parskip}
  }{% else
    \setlength{\parindent}{0pt}
    \setlength{\parskip}{6pt plus 2pt minus 1pt}}
}{% if KOMA class
  \KOMAoptions{parskip=half}}
\makeatother
\usepackage{xcolor}
\usepackage[margin=1in]{geometry}
\usepackage{color}
\usepackage{fancyvrb}

\DefineVerbatimEnvironment{Highlighting}{Verbatim}{commandchars=\\\{\}}
\newenvironment{Shaded}{}{}

\newcommand{\DataTypeTok}[1]{\textcolor[rgb]{0.56,0.13,0.00}{#1}}

\newcommand{\ErrorTok}[1]{\textcolor[rgb]{1.00,0.00,0.00}{\textbf{#1}}}

\newcommand{\FloatTok}[1]{\textcolor[rgb]{0.25,0.63,0.44}{#1}}
\newcommand{\FunctionTok}[1]{\textcolor[rgb]{0.02,0.16,0.49}{#1}}

\newcommand{\OtherTok}[1]{\textcolor[rgb]{0.00,0.44,0.13}{#1}}

\newcommand{\StringTok}[1]{\textcolor[rgb]{0.25,0.44,0.63}{#1}}

\usepackage{longtable,booktabs,array}
\usepackage{calc} % for calculating minipage widths
\usepackage{etoolbox}
\makeatletter
\patchcmd\longtable{\par}{\if@noskipsec\mbox{}\fi\par}{}{}
\makeatother
\IfFileExists{footnotehyper.sty}{\usepackage{footnotehyper}}{\usepackage{footnote}}
\makesavenoteenv{longtable}
\providecommand{\tightlist}{%
  \setlength{\itemsep}{0pt}\setlength{\parskip}{0pt}}
\usepackage{tikz}
\usetikzlibrary{arrows.meta,positioning,shapes.geometric}
\usepackage{xurl}  % allow long URLs (GitHub/Zenodo) to break across lines
\ifLuaTeX
  \usepackage{selnolig}  % disable illegal ligatures
\fi
\usepackage[numbers,sort&compress]{natbib}
\IfFileExists{bookmark.sty}{\usepackage{bookmark}}{\usepackage{hyperref}}
\IfFileExists{xurl.sty}{\usepackage{xurl}}{} % add URL line breaks if available
\hypersetup{
  pdftitle={Power Systems Agent Benchmark: Executable Evaluation of AI Agents in Electric Power Engineering},
  pdfauthor={Sergei Trashchenkov},
  colorlinks=true,
  linkcolor={black},
  filecolor={Maroon},
  citecolor={blue},
  urlcolor={blue},
  pdfcreator={LaTeX via pandoc}}

\title{Power Systems Agent Benchmark: Executable Evaluation of AI Agents
in Electric Power Engineering}
\author{Sergei Trashchenkov\footnote{trashchenkov@gmail.com; ORCID
  0000-0001-8786-8336}}
\date{}

\begin{document}
\maketitle
\begin{abstract}
Executable evaluation --- having an agent act and then checking the
consequences of that action with a program rather than grading its prose
--- has become a prominent way to assess tool-using AI agents in
software and related settings. Electric power engineering has not yet
had an analogous benchmark: language-model use in the domain is still
dominated by retrieval and text question answering, while agents that
act on power-system artifacts remain mostly academic prototypes. We
introduce the Power Systems Agent Benchmark, an executable benchmark for
power-engineering agents. An agent receives a structured task and
returns a structured solution; a deterministic evaluator then recomputes
the relevant engineering quantities, checks operational constraints, and
returns a feasibility flag, a normalized score, and explicit violations.

The benchmark contains 41 task families across eight areas of power
engineering, including power flow, protection, stability and grid-code
compliance, distributed resources, microgrids, reliability, power
quality, and forecasting. Each task is grounded in a citable source,
standard, textbook method, or documented engineering formulation. To
reduce contamination, held-out cases are synthesized on demand by
per-family generators from private seeds: the construction is
inspectable, but the concrete instances remain private. In a reference
evaluation with three command-line agents, the strongest configurations
score near the compact tier's ceiling, a smaller open-model
configuration trails, and public and held-out performance are broadly
consistent. We also report a separate public-split model-harness grid
with OpenCode and Aider. The reference evaluation doubles as a
quality-control procedure: unanimous failures flag candidate task or
evaluator defects, and it exposed a latent evaluator bug missed by
self-consistency checks. The current evaluators are compact
deterministic surrogates, but the task contract allows selected
internals to be upgraded to simulator-backed checks without changing how
tasks are posed or solved.
\end{abstract}

\subsection{1. Introduction}\label{introduction}

Large language models have, over a short period, moved from systems that
produce text to systems that take actions. Given access to tools, a
model can read structured data, call an external program, inspect the
result, and decide what to do next; wrapped in such a loop, it becomes
what is now commonly called an agent. This shift has been most visible
in software, where agents resolve issues in real code repositories, and
in business operations, where they handle customer workflows, query
databases, and execute multi-step procedures through tool and API calls.
The appeal is the same across these settings: an agent can carry out
work that previously required a person to translate intent into a
sequence of tool invocations.

It is natural to ask whether the same capability transfers to
specialized engineering domains, where the work is governed not by
conventions but by physical law. Electric power systems are a demanding
instance. Engineering decisions there --- estimating a fault current,
choosing a protective relay setting, scheduling generation under
security constraints, screening a network for contingencies --- rest on
circuit theory, on stability and protection principles, and on a dense
body of standards. Whether language-model agents can perform such work
is an open research question, and an active one: recent surveys document
a rapid influx of agent and LLM methods into power-system analysis
\citep{llmpowersurvey2025, agenticpowersurvey2025}, and early prototype
systems have begun to appear \citep{gridmind2025, xgridagent2025}. This
activity is, for now, almost entirely experimental rather than deployed,
which makes the question of how to measure progress a pressing one in
its own right.

Measuring it is harder than it first appears, because correctness in
power engineering is a physical property, not a textual one. A proposed
relay setting can be eloquently justified and fail to be selective; a
dispatch can be internally consistent and exceed a generator's limits; a
restoration plan can read well and de-energize the wrong feeder. Errors
of this kind are invisible to a human skimming the explanation and to a
language model asked to judge it, because they are not errors of
expression --- they are violations of network physics and operating
limits, and they surface only when the proposed solution is recomputed
against the case. The two evaluation methods most common in agent
research are, for precisely this reason, ill-suited here: question
answering against a fixed reference answer cannot accommodate the many
admissible solutions an engineering task allows, and grading a free-form
report with a second model \citep{zheng2023judging} rewards a fluent
rationale even when the underlying decision is unsound --- a tendency
reinforced by the documented biases of such judges
\citep{wang2023fairevaluators}.

In software and tool-use settings this difficulty has been met by
executable evaluation: an agent's output is not scored as prose but
applied and checked by a program. SWE-bench \citep{jimenez2024swebench}
runs a repository's own test suite against the agent's patch;
\(\tau\)-bench \citep{yao2024taubench} compares the final state of a
database against an annotated goal; terminal-bench
\citep{terminalbench2025} inspects the effect of an agent's commands on
a live shell. The principle these share --- judge the outcome of an
action, not its description --- is exactly what power-engineering
evaluation needs, and it has not been applied to the domain at any
breadth.

This paper does so. We introduce the Power Systems Agent Benchmark, in
which each task supplies an agent with structured inputs and a required
output schema; the agent returns a structured solution; and a
deterministic evaluator recomputes the relevant engineering quantities,
tests the hard constraints, and returns a feasibility flag, a normalized
score in the unit interval, and an explicit list of violations. The
agent's natural-language explanation is recorded but excluded from
scoring --- this keeps explanations useful for error analysis without
letting a fluent but physically invalid rationale raise the score. The
public portion comprises 41 task families across eight areas of power
engineering, each grounded in a citable source, standard, textbook
method, or documented engineering formulation. The held-out portion is
not distributed as data: a generator produces held-out cases on demand
from a private seed, so that the construction of the held-out set is
open to inspection while its instances remain unknown to those being
evaluated.

This benchmark is our primary contribution, and it is best understood as
three things at once. It is, first, a broad executable test of
power-engineering competence: 41 task families spanning eight areas of
the field, each posed through a uniform contract and checked by a
deterministic evaluator that recomputes the underlying physics, so that
an agent is measured by what its decisions do rather than by how it
describes them. It is, second, a contamination-resistant evaluation:
rather than ship a fixed held-out set that ages and leaks, we provide a
per-family generator that synthesizes physically consistent,
solvability-checked cases from a private seed, keeping the held-out
construction auditable while its instances stay private. And it is,
third, an agent-agnostic platform: because participating reduces to
reading a task file and writing a solution file, the same benchmark can
compare different language models and different agent harnesses under
one protocol, and can serve as a substrate for studying agent behavior
on physically grounded tasks. We accompany the benchmark with a
reference evaluation of three command-line agents and, in the course of
running it, report a methodological observation --- that a unanimous
agent failure is a useful signal of a defective task or evaluator ---
strong enough to recommend as a quality-control step in its own right
(Section 8.2). A separate public-split model-harness grid with OpenCode
and Aider then illustrates how the same benchmark can separate model and
scaffolding effects (Section 8.4). Evaluation here is not only for
ranking frontier agents: it also serves as a deployment filter for the
cheaper, local, or open models that practitioners try first, which
Section 8.4 shows still err substantially. We frame this as a v0: a
broad, contamination-resistant, agent-agnostic executable prototype
across 41 compact task families --- not a claim of priority over earlier
domain efforts (Section 2), nor a measure of industrial-scale competence
(Section 9).

\subsection{2. Related Work}\label{related-work}

Our work draws on three lines of research, and is most easily located by
where it follows each and where it departs.

The first is executable evaluation of agents. SWE-bench
\citep{jimenez2024swebench} and the earlier pass@k construction of
Codex/HumanEval \citep{chen2021codex} established that an agent's output
can be judged by executing it; \(\tau\)-bench \citep{yao2024taubench}
generalized this to tool-and-user interaction by comparing end states
and proposed pass\^{}k as a reliability measure; terminal-bench
\citep{terminalbench2025} and MLE-bench \citep{chan2024mlebench} extend
the idea to terminal work and machine-learning engineering, while broad
suites such as AgentBench \citep{liu2024agentbench}, GAIA
\citep{mialon2023gaia}, and WebArena \citep{zhou2024webarena} probe
general tool use. We adopt this paradigm; our point of difference is
coverage, since none of these targets power-system engineering and its
physical constraints.

The second line concerns \emph{how} to score, and it is the reason we
recompute rather than judge. What ultimately matters is the effect of an
agent's actions, not the fluency of its account of them, and for
verifying effects the strongest available tool is a deterministic check.
This is not a new idea --- it is the basis of software testing, and of
competitive programming, where a submission is accepted only if it
passes a fixed suite of tests --- and wherever a task admits such a
check, as every task in this benchmark does, it should be preferred: it
is exact, reproducible, and immune to the biases that attend a
language-model judge, the biases of position, length, and
self-preference \citep{zheng2023judging, wang2023fairevaluators} that
are most damaging exactly when a wrong answer is fluent. Model-based
judging retains a role where no deterministic check is feasible, and
recent work extends it from single responses to whole agent trajectories
under the heading of \emph{agent-as-a-judge}
\citep{zhuge2024agentjudge, aisjudgeais2025}. We expect that as the
benchmark grows to include scenarios too open-ended for a closed-form
evaluator, judging of this kind --- applied to actions and trajectories,
not to prose --- will become appropriate, and we leave that path open;
for the tasks presented here, deterministic recomputation suffices, and
the agent's rationale is kept only as a diagnostic.

The third line is domain-specific. Surveys
\citep{llmpowersurvey2025, agenticpowersurvey2025} record a rapid growth
of LLM and agent methods in power systems with no settled means of
evaluation. PowerAgentBench \citep{poweragentbench}, part of the
actively developed PowerAgent ecosystem at Harvard SEAS
\citep{zhang2025poweragent}, is the closest prior effort and already
instantiates the public/held-out, evaluator-based design in this domain.
Our contribution is complementary: broader coverage across task
families, seed-generated held-out cases, an agent-agnostic command-line
protocol, and an explicit quality-control procedure for detecting task
and evaluator defects. ElecBench \citep{zhou2025elecbench} scores
dispatch scenarios largely through text and judges, and PFBench
\citep{pfbench2025} applies programmatic checking but only to power
flow. The prototype agents now appearing --- GridMind
\citep{gridmind2025}, X-GridAgent \citep{xgridagent2025}, Grid-Mind CIA
\citep{gridmindcia2026}, PFAgent \citep{pfagent2026}, and related
systems --- are the intended subjects of a benchmark of this kind.

Two further bodies of work inform specific choices.
Reinforcement-learning grid environments, principally Grid2Op/L2RPN
\citep{marot2020l2rpn, marot2021retrospective}, provide a mature setting
for sequential topology control and are built on established
power-system simulators --- pandapower \citep{thurner2018pandapower},
MATPOWER \citep{zimmerman2011matpower}, OpenDSS
\citep{dugan2011opendss}, PowerModels/PGLib
\citep{coffrin2018powermodels, babaeinejadsarookolaee2019pglib}, ANDES
\citep{cui2021andes}, PyPSA \citep{brown2018pypsa} --- refined over many
years to compute power-system physics. We address the different problem
of single-shot engineering decisions across many subtasks under an
agent-agnostic contract, and our current evaluators do not use these
simulators at all: each is a compact, self-contained deterministic
check. We name them only as the natural backend for higher-fidelity
future versions of the evaluators, not as a present dependency. Finally,
the literature on benchmark integrity informs the held-out split:
because training-data contamination inflates results
\citep{contaminationsurvey2025, statictodynamic2025}, dynamic and
refreshed benchmarks have become the standard response
\citep{kiela2021dynabench, white2024livebench, jain2024livecodebench},
and a seed-driven generator is a prevention mechanism of the same
family. The runner follows the prescriptions of
\citep{kapoor2024agentsmatter} on cost, reproducibility, and the
separation of failure classes, and respects the limits of best-of-N
characterized in \citep{brown2024monkeys}.

\subsection{3. Design Requirements}\label{design-requirements}

The system is shaped by a small set of requirements, stated here at the
outset because most of what follows is a consequence of them; we note
for each where it is realized.

\begin{enumerate}
\def\labelenumi{\arabic{enumi}.}
\tightlist
\item
  \textbf{Executable, physical verification.} A solution must be checked
  by running code rather than by matching strings, and that code must
  recompute the relevant physical quantities and test the operating
  constraints, not compare against stored text. The remaining
  requirements largely follow from this one (the contract is given in
  Section 4.1, the evaluators in Section 6).
\item
  \textbf{A leak-proof public/held-out split.} Public cases must support
  development and held-out cases a fair comparison; crucially, held-out
  answers must be excluded by construction rather than by policy, so
  that no amount of scrutiny of the released material reveals them
  (Section 4.4).
\item
  \textbf{Agent agnosticism.} Since the point is to compare models and
  harnesses on equal terms, launching an agent must not depend on any
  particular one --- participation reduces to reading a task file and
  writing a solution file (Section 4.2).
\item
  \textbf{Provenance and machine-readable results.} Every task must
  declare where it comes from, with a confidence level (Section 5.1),
  and every result must be machine-readable --- a feasibility flag, a
  score, and an explicit list of violations (Sections 4.1 and 6) --- so
  that runs can be aggregated and audited without anyone reading prose.
\item
  \textbf{Room to grow.} The benchmark must be able to become more
  demanding without breaking compatibility, so that replacing a
  closed-form check with a simulator-backed one changes an evaluator's
  internals while leaving the task and solution formats untouched
  (Section 4.3).
\item
  \textbf{Specification-quality control.} Finally, because a benchmark's
  own tasks and evaluators can themselves be wrong, it must provide a
  means of telling a defect in its specifications apart from a genuine
  difference in agent capability (demonstrated in Section 8.2).
\end{enumerate}

\subsection{4. Architecture}\label{architecture}

\subsubsection{4.1 Artifact contract}\label{artifact-contract}

The benchmark is organized around a single linear contract (Figure 1): a
task description is given to an agent, the agent produces a structured
solution, an evaluator consumes the solution, and an evaluation result
is emitted. A JSON schema fixes each of the three artifacts. A task
carries its identity and family, its public inputs, the schema its
solution must follow, the hard constraints that will be checked, the
metrics, the provenance, and a bilingual presentation block. A solution
carries the agent's structured answer and an optional rationale. An
evaluation result carries the feasibility flag, the score, the list of
violations, and any supporting evidence. Because the formats are fixed
independently of how a task is solved or checked, an agent and an
evaluator need share nothing beyond these schemas.

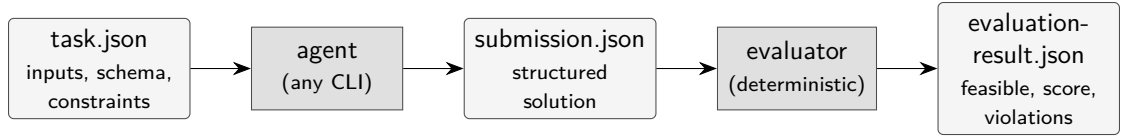
\begin{figure}[t]
\centering
\begin{tikzpicture}[
  font=\sffamily\small,
  artifact/.style={rectangle, rounded corners=2pt, draw=black!70, fill=black!4,
                   minimum height=11mm, minimum width=24mm, align=center, inner sep=4pt},
  actor/.style={rectangle, draw=black!70, fill=black!12, minimum height=11mm,
                minimum width=20mm, align=center, inner sep=4pt},
  >={Stealth[length=2.4mm]}, node distance=8mm]
  \node[artifact] (task) {task.json\\\scriptsize inputs, schema,\\[-1pt]\scriptsize constraints};
  \node[actor, right=of task] (agent) {agent\\\scriptsize (any CLI)};
  \node[artifact, right=of agent] (sub) {submission.json\\\scriptsize structured\\[-1pt]\scriptsize solution};
  \node[actor, right=of sub] (eval) {evaluator\\\scriptsize (deterministic)};
  \node[artifact, right=of eval] (res) {evaluation-\\result.json\\\scriptsize feasible, score,\\[-1pt]\scriptsize violations};
  \draw[->] (task) -- (agent);
  \draw[->] (agent) -- (sub);
  \draw[->] (sub) -- (eval);
  \draw[->] (eval) -- (res);
\end{tikzpicture}
\caption{The benchmark contract. A task is posed to any command-line agent, which writes a structured solution; a deterministic evaluator recomputes the result and emits a machine-readable verdict.}
\label{fig:contract}
\end{figure}

A short-circuit task, abbreviated, illustrates the form:

\begin{Shaded}
\begin{Highlighting}[]
\FunctionTok{\{}
  \DataTypeTok{"family"}\FunctionTok{:} \StringTok{"three\_phase\_short\_circuit"}\FunctionTok{,}
  \DataTypeTok{"public\_inputs"}\FunctionTok{:} \FunctionTok{\{}\DataTypeTok{"short\_circuit\_case"}\FunctionTok{:} \FunctionTok{\{}
    \DataTypeTok{"base\_mva"}\FunctionTok{:} \FloatTok{100.0}\FunctionTok{,} \DataTypeTok{"voltage\_kv"}\FunctionTok{:} \FloatTok{10.5}\FunctionTok{,}
    \DataTypeTok{"voltage\_factor\_cmax"}\FunctionTok{:} \FloatTok{1.1}\FunctionTok{,} \DataTypeTok{"voltage\_factor\_cmin"}\FunctionTok{:} \FloatTok{1.0}\FunctionTok{,}
    \DataTypeTok{"positive\_sequence\_impedances\_pu"}\FunctionTok{:} \OtherTok{[} \ErrorTok{/*} \ErrorTok{...} \ErrorTok{*/} \OtherTok{]}\FunctionTok{,}
    \DataTypeTok{"fault\_resistance\_pu"}\FunctionTok{:} \FloatTok{0.005}\FunctionTok{\}\},}
  \DataTypeTok{"submission\_schema"}\FunctionTok{:} \FunctionTok{\{}
    \DataTypeTok{"ik\_max\_ka"}\FunctionTok{:} \StringTok{"max initial symmetrical current (kA)"}\FunctionTok{,}
    \DataTypeTok{"ik\_min\_ka"}\FunctionTok{:} \StringTok{"min current with fault resistance (kA)"}\FunctionTok{,} \DataTypeTok{"..."}\FunctionTok{:} \StringTok{"..."}\FunctionTok{\},}
  \DataTypeTok{"hard\_constraints"}\FunctionTok{:} \OtherTok{[}
    \StringTok{"maximum current uses c\_max over a bolted fault"}\OtherTok{,}
    \StringTok{"values match the independent positive{-}sequence calculation"}\OtherTok{]}\FunctionTok{,}
  \DataTypeTok{"metrics"}\FunctionTok{:} \OtherTok{[}\FunctionTok{\{}\DataTypeTok{"name"}\FunctionTok{:} \StringTok{"feasible"}\FunctionTok{,} \DataTypeTok{"direction"}\FunctionTok{:} \StringTok{"boolean"}\FunctionTok{,} \DataTypeTok{"..."}\FunctionTok{:} \StringTok{"..."}\FunctionTok{\}}\OtherTok{,}
              \FunctionTok{\{}\DataTypeTok{"name"}\FunctionTok{:} \StringTok{"score"}\FunctionTok{,} \DataTypeTok{"direction"}\FunctionTok{:} \StringTok{"maximize"}\FunctionTok{,} \DataTypeTok{"..."}\FunctionTok{:} \StringTok{"..."}\FunctionTok{\}}\OtherTok{]}\FunctionTok{,}
  \DataTypeTok{"source\_provenance"}\FunctionTok{:} \OtherTok{[}\FunctionTok{\{}\DataTypeTok{"title"}\FunctionTok{:} \StringTok{"IEC 60909 short{-}circuit calculation"}\FunctionTok{,}
                         \DataTypeTok{"confidence"}\FunctionTok{:} \StringTok{"high"}\FunctionTok{,} \DataTypeTok{"..."}\FunctionTok{:} \StringTok{"..."}\FunctionTok{\}}\OtherTok{]}
\FunctionTok{\}}
\end{Highlighting}
\end{Shaded}

\subsubsection{4.2 Agent-agnostic execution and a trustworthy
runner}\label{agent-agnostic-execution-and-a-trustworthy-runner}

For a comparison across agents to be meaningful, the \emph{runner} ---
the benchmark's own apparatus that launches each agent (itself a harness
of model, loop, and tools) on each task and scores what it returns ---
must satisfy three properties beyond merely invoking the agent. The
first is isolation: an agent must not be able to read the evaluator or
any stored answer. Each task is therefore packaged with its answer key
and evaluator internals removed, and is executed in a sandbox that has
no path back to the benchmark's own code. This matters because an agent
with autonomous filesystem access can, in principle, reach an evaluator
that is within reach and grade itself against it; a written instruction
not to do so is no safeguard, since nothing compels the agent to honor
it. Isolation must therefore be structural rather than instructional.

The second property is robustness to malformed output. An agent may
return a solution that is syntactically valid but meaningless --- null
values, text where a number is required, the wrong structure entirely
--- and the evaluator must treat this as an ordinary failed solution,
returning infeasible with a list of the offending fields, rather than
raising an exception that halts the entire run. Without this property a
single malformed solution can abort a multi-task evaluation partway and
leave an incomplete, incomparable record.

The third property is a principled separation of failure classes. A run
can fail for reasons that have nothing to do with the agent's competence
--- a transient provider error, a dropped connection, an exhausted usage
quota --- and conflating these with genuine failures contaminates the
measurement. The runner classifies each failure and retries only those
attributable to infrastructure, never those in which the agent produced
a poor or absent solution within a healthy environment; retrying the
latter would silently turn the benchmark into a best-of-N search and
inflate every score. Each result records how many attempts it took and
why earlier attempts failed, and a run reports its metrics both with and
without infrastructure retries, so that provider reliability is never
mistaken for agent capability.

\subsubsection{4.3 Evaluator maturity
ladder}\label{evaluator-maturity-ladder}

The evaluators are allowed to mature in place along a ladder: from a
closed-form or graph-based surrogate, through a deterministic
engineering model, to a check backed by a full power-system simulator,
and ultimately to a stochastic or large-scale scenario suite. The
evaluators presented here occupy the first two rungs, and within those
rungs they are of two kinds. About half compute an engineering quantity
in closed form from a standard or textbook method --- short-circuit and
symmetrical-component fault currents, an admittance matrix, a
transformer thermal model, harmonic and voltage-compliance limits ---
and are exact for the compact problem as posed. The rest score a
decision over a deliberately simplified model --- a linear
voltage-sensitivity stand-in for an AC power flow, or an optimization
over a compact dispatch or restoration model --- and are approximate by
construction; the boundary is not always sharp, since some solve a
linearized model exactly. None yet invokes a full power-system
simulator, which is the next rung. The contract of Section 4.1 is what
makes the ladder usable: because the task and solution formats are fixed
independently of how a check is computed, the internal step of
replacing, say, a linearized power-flow surrogate with a full AC
solution upgrades an evaluator without altering anything an agent sees.
The ladder describes how a check is computed, but the contract is
symmetric in the tools it admits: nothing stops an agent, on its side,
from invoking the same kind of software to produce a solution --- a
possibility today's tasks do not yet exercise but later ones will
(Section 10).

\subsubsection{4.4 The held-out split as a
generator}\label{the-held-out-split-as-a-generator}

The benchmark does not store its held-out cases at all. In place of
held-out data files, it provides a generator: a collection of
deterministic, physics-aware perturbations, one per family. A held-out
case is produced from a public case by redrawing its numerical content
--- network parameters, weather series, load profiles, a fault location
together with the indicator readings and customer complaints it implies
--- under a private random seed, and a draw is accepted only if the
family's reference solver still solves it, the evaluator confirms the
solution feasible, and no small perturbation of that solution scores
higher, a local check on the reference solution used for scoring
(Section 6), with a deterministic redraw otherwise; Section 8.3 confirms
empirically that this yields solvable, diverse held-out instances at
scale. In a leaderboard setting, the operator keeps the seed private,
while anyone else can read exactly how held-out cases are constructed
without learning the particular instances of any round. This makes the
held-out set, normally the most leak-prone component of a benchmark,
auditable in mechanism and private in content at the same time. The same
discipline extends to the records of evaluation runs: an agent's
solution to a held-out case, and the evaluator's per-case output, each
disclose that case's answer, so only aggregate run summaries are
published --- never the per-case held-out submissions, results, or logs.

\subsection{5. Task Set}\label{task-set}

The public benchmark comprises 41 task families, grouped into eight
domain areas:

\begin{enumerate}
\def\labelenumi{\arabic{enumi}.}
\tightlist
\item
  \textbf{Network analysis and operating constraints} ---
  bus-admittance-matrix construction, corrective reactive-power dispatch
  for voltage limits, contingency screening \citep{zhou2019contingency},
  post-contingency corrective switching, dynamic line rating
  \citep{ieee738}.
\item
  \textbf{Short circuit and protection} --- three-phase and
  single-line-to-ground fault calculation
  \citep{iec60909, anderson1995faulted}, breaker and relay setting from
  fault levels, distance-protection zone setting and overcurrent
  coordination \citep{horowitz2014relaying}, cable sizing
  \citep{iec60364552}.
\item
  \textbf{Stability, grid code, and inverter-based resources} ---
  critical clearing time by the equal-area criterion
  \citep{kundur1994stability}, transient-stability classification,
  fault-ride-through compliance \citep{ieee2800, entsoerfg},
  current-limited inverter fault response, minimum
  synchronous-generation share under inertia and short-circuit
  constraints.
\item
  \textbf{Distributed resources, PV, EV, and storage} --- Volt-VAR
  control \citep{turitsyn2010voltvar}, vehicle-to-grid scheduling and
  voltage support, battery ancillary-service response, soft-cost
  levelized-cost calculation.
\item
  \textbf{Microgrids and dispatch} --- economic and rolling-horizon
  dispatch, islanded active/reactive dispatch, robust dispatch under
  uncertainty, hydro-thermal unit commitment, single-contingency-secure
  generation commitment \citep{scuccts2020}.
\item
  \textbf{Reliability and restoration} --- fault isolation and service
  restoration \citep{restoration2020}, faulted-section localization,
  fault-indicator placement, reliability-index improvement
  \citep{ieee1366}, operator switching and shedding.
\item
  \textbf{Power quality, standards, assets, and cybersecurity} ---
  voltage-compliance checking to a supply standard \citep{en50160},
  power-quality event classification, harmonic-limit compliance
  \citep{ieee519}, transformer thermal loading and insulation ageing
  \citep{iec600767}, bad-data identification in state estimation
  \citep{abur2004se}, secured-measurement placement \citep{liu2011fdi},
  lead-station handover in a multi-agent frequency scheme.
\item
  \textbf{Forecasting under uncertainty} --- wind power forecasting,
  prediction-interval estimation.
\end{enumerate}

Each task was checked for physical correctness against its cited source
before inclusion. Every family carries a documented source and
confidence level (Section 5.1), and the appendix lists the full catalog
with identifiers.

\subsubsection{5.1 Provenance and
confidence}\label{provenance-and-confidence}

Because the task set is drawn from the engineering literature rather
than authored from scratch, each family records where its formulation
comes from and how firmly it is tied to that origin. Every task carries
a provenance entry --- a source title, a locator (a digital object
identifier or URL for an article, a clause for a standard, or a full
citation for a textbook), and a short note on how the source maps onto
the executable case --- together with a stated confidence level. The
confidence level is a deliberate, declared judgment rather than a
measured quantity, and we assign it by a fixed rubric so that it is
reproducible:

\begin{itemize}
\tightlist
\item
  \textbf{high} --- the task reproduces a method defined in a published
  standard or a canonical textbook, cited specifically, and the
  executable case follows that method directly (for example, the
  short-circuit \citep{iec60909}, transformer-loading \citep{iec600767},
  and harmonic-compliance \citep{ieee519} tasks follow named
  international standards, and the critical-clearing-time
  \citep{kundur1994stability} and bad-data \citep{abur2004se} tasks
  follow standard textbook procedures);
\item
  \textbf{medium} --- the task is tied to an identified primary source,
  or to a universally taught engineering method written out as its
  governing equations, and the case is a faithful compact instance of
  it;
\item
  \textbf{low} --- no single primary document is pinned: the method is
  standard but the task is an adaptation in its spirit rather than a
  reproduction of a specific source.
\end{itemize}

High here does not mean industrial validation; it means tight
traceability to a standard or canonical textbook procedure. Under this
rubric, of the 41 families 11 are rated high, 28 medium, and 2 low; the
two low cases (uncertainty-aware dispatch and minimum synchronous share)
keep that rating because no single source we could verify fully grounds
their exact formulation, and we prefer to say so than to inflate the
label. We regard the scheme as a floor on transparency rather than a
guarantee of fidelity: it lets a reader trace any task to its origin and
see at a glance how tightly it is bound there, and it makes explicit
which tasks are faithful renderings of a standard and which are looser
adaptations of common practice.

\subsection{6. Evaluators and Metrics}\label{evaluators-and-metrics}

Every evaluation returns the same four things: a feasibility flag, a
score in the unit interval, a list of violations stated in words, and
task-specific evidence. The two numbers play distinct roles. Each task
declares a set of \emph{hard constraints} --- the physical and
operational requirements a solution must satisfy to be admissible at
all, such as respecting a generator's limits, holding every voltage
within its band, or keeping the network radial --- and a solution is
\emph{feasible} only if it violates none of them. The score then grades
quality among feasible solutions; an infeasible solution scores zero
however close it may have looked.

How the score is computed depends on the task, but every score is
reported \emph{relative to the reference optimum}: it is scaled so that
an optimal solution earns exactly one, and a feasible but worse solution
earns a fraction below it. This matters because many tasks have no
natural unit ceiling on their raw objective --- a reactive effort in
kvar, a switching cost, an amount of restored load --- so without this
scaling even a perfect answer would score below one. The reference
solver for each family computes that optimum --- exact where the task is
closed-form, and via the family's optimization procedure otherwise (with
the caveats of Section 9); an agent's solution, being one feasible
solution among many, cannot exceed it, and should a submission
nonetheless score higher the reference was not optimal on that case and
the reported score is capped at one. An exact computation is scored
against numeric tolerances --- a 0.1\% relative band with a small
absolute floor, so that an answer correct to about four significant
figures passes while a genuinely wrong one does not; a control or
optimization task scores the gap between the submitted objective and the
optimum, so that an alternative optimum of equal cost receives full
credit and a feasible but costlier solution receives partial credit; a
classification or localization task is scored against exact sets with
enumerated misses; and a standards-compliance task as pass/fail with
diagnostic fields.

A microgrid dispatch task makes the distinction concrete: its hard
constraints are the power balance, the generator and battery limits, and
the state-of-charge bounds; any schedule meeting all of them is
feasible, and among feasible schedules the least-cost schedule scores
exactly one while a costlier feasible dispatch scores below one in
proportion to its cost gap.

At the level of a whole run we report two aggregates that need not
agree: the \emph{feasible rate}, the fraction of tasks whose solution
satisfied every hard constraint, and the \emph{perfect rate}, the
fraction scoring exactly one (feasible and, where graded, optimal). The
two diverge whenever an agent produces admissible but suboptimal
solutions, and because they can rank agents differently we report both
alongside the mean score.

\subsection{7. Reproducibility}\label{reproducibility}

The benchmark repository is public at
\url{https://github.com/trashchenkov/power-systems-agent-benchmark}; the
paper artifact release is archived on Zenodo with DOI
\url{https://doi.org/10.5281/zenodo.20753046}. Every number in this
paper can be regenerated from the repository. A single command runs the
unit tests over the evaluators, the reference solvers, the runner, and
the generator; validates every artifact; evaluates each of the 41 public
sample solutions individually; and runs the reference solvers across the
public set, which return all 41 cases feasible. Because each task's
score is measured relative to its own reference optimum (Section 6), a
reference solution scores one by construction, so the check confirms not
that the tasks are easy but that every task has a contract-valid,
physically consistent reference solution. One limit of such a check is
worth stating plainly: a reference solver and an evaluator built on the
same kernels can share the same mistake and still agree, so their
agreement confirms the contract but cannot, by itself, certify the
evaluator's underlying physics. As a partial, independent check on that
physics, a set of evaluators is additionally cross-checked against
golden cases whose answers are derived by hand from the governing
formulas rather than from our own baselines --- at present three:
three-phase short circuit (IEC 60909), single-line-to-ground fault by
symmetrical components, and admittance-matrix construction. Continuous
integration runs the full check on every change.

\subsection{8. Experiments}\label{experiments}

We evaluated three command-line agents on an identical task input ---
the same \texttt{task.json} and the same instruction file for every
agent on every task --- under a fixed per-task timeout and the
infrastructure-retry policy of Section 4.2, with no adaptation of the
agents to the benchmark. Each agent was invoked through its own
command-line interface and ran under whatever internal system prompt and
scaffolding it ships with; we standardize what the task asks, not how an
agent organizes itself to answer it, since that scaffolding is part of
what an agent is and what the benchmark compares. The three are Codex
(codex-cli 0.139.0) running gpt-5.5 at medium reasoning effort, Cursor's
command-line agent (cursor-agent 2026.06.12) running composer-2.5, and
the open-source OpenCode agent (opencode 1.15.13) running
deepseek-v4-flash-free, the free tier of that model through OpenCode's
provider; we refer to them below as Codex, Cursor, and OpenCode. Both
the models and these agent CLIs are updated frequently, so the round,
run in June 2026, should be read as a snapshot of these specific
versions rather than a standing ranking; re-running the same protocol
with later releases may give different results. This reference round is
the subject of Sections 8.1 and 8.2; Section 8.3 then validates the
held-out generator and Section 8.4 widens the comparison beyond these
three to further models and a second harness.

\subsubsection{8.1 Public versus held-out
evaluation}\label{public-versus-held-out-evaluation}

Each agent attempted the public case of every family and one freshly
generated held-out case per family from a new private seed, every
solution scored by the same deterministic, optimum-normalized
evaluators.

\begin{longtable}[]{@{}
  >{\raggedright\arraybackslash}p{(\columnwidth - 12\tabcolsep) * \real{0.1429}}
  >{\raggedright\arraybackslash}p{(\columnwidth - 12\tabcolsep) * \real{0.1429}}
  >{\raggedright\arraybackslash}p{(\columnwidth - 12\tabcolsep) * \real{0.1429}}
  >{\raggedright\arraybackslash}p{(\columnwidth - 12\tabcolsep) * \real{0.1429}}
  >{\raggedright\arraybackslash}p{(\columnwidth - 12\tabcolsep) * \real{0.1429}}
  >{\raggedright\arraybackslash}p{(\columnwidth - 12\tabcolsep) * \real{0.1429}}
  >{\raggedright\arraybackslash}p{(\columnwidth - 12\tabcolsep) * \real{0.1429}}@{}}
\caption{Public and held-out performance of the three reference agents
(feasible, perfect, and mean score out of 41). Each agent bundles a
model with its harness; Section 8.4 separates the two
effects.}\tabularnewline
\toprule\noalign{}
\begin{minipage}[b]{\linewidth}\raggedright
agent
\end{minipage} & \begin{minipage}[b]{\linewidth}\raggedright
public feasible
\end{minipage} & \begin{minipage}[b]{\linewidth}\raggedright
public perfect
\end{minipage} & \begin{minipage}[b]{\linewidth}\raggedright
public mean
\end{minipage} & \begin{minipage}[b]{\linewidth}\raggedright
held-out feasible
\end{minipage} & \begin{minipage}[b]{\linewidth}\raggedright
held-out perfect
\end{minipage} & \begin{minipage}[b]{\linewidth}\raggedright
held-out mean
\end{minipage} \\
\midrule\noalign{}
\endfirsthead
\toprule\noalign{}
\begin{minipage}[b]{\linewidth}\raggedright
agent
\end{minipage} & \begin{minipage}[b]{\linewidth}\raggedright
public feasible
\end{minipage} & \begin{minipage}[b]{\linewidth}\raggedright
public perfect
\end{minipage} & \begin{minipage}[b]{\linewidth}\raggedright
public mean
\end{minipage} & \begin{minipage}[b]{\linewidth}\raggedright
held-out feasible
\end{minipage} & \begin{minipage}[b]{\linewidth}\raggedright
held-out perfect
\end{minipage} & \begin{minipage}[b]{\linewidth}\raggedright
held-out mean
\end{minipage} \\
\midrule\noalign{}
\endhead
\bottomrule\noalign{}
\endlastfoot
Codex & 40/41 & 38/41 & 0.973 & 41/41 & 40/41 & 0.995 \\
Cursor & 40/41 & 39/41 & 0.973 & 41/41 & 41/41 & 1.000 \\
OpenCode & 39/41 & 39/41 & 0.951 & 40/41 & 39/41 & 0.973 \\
\end{longtable}

Two things stand out, and the first is sobering for a benchmark: the
strongest agents score so close to the maximum that it has little room
left to tell them apart. Codex and Cursor are nearly indistinguishable
on the public split and essentially solve the held-out one, and OpenCode
trails only modestly. We therefore do not present this round as a sharp
ranking of frontier agents; on this compact tier the benchmark has
limited power to separate the best of them, a point we take up in
Section 9.

Second, the two splits track each other --- every agent scores at least
as high on the freshly generated held-out cases as on the public ones.
The signature one would expect from contamination, public scores
standing above held-out, is absent; if anything the held-out draws were
marginally easier. We make no claim of an intrinsic difficulty gap
between the splits, and the case for the generator is prospective: a
published set ages as models train on it, and the generator is what
keeps an uncontaminated draw available.

What discrimination the benchmark does provide comes at the lower end of
the capability range and through shared blind spots, not through margins
among the leaders: OpenCode fails a microgrid-dispatch task outright
where the others succeed, and one storage-scheduling task is missed by
every agent in the same way --- a genuine capability gap, confirmed not
to be a specification defect (Section 8.2).

\subsubsection{8.2 Distinguishing capability from
specification}\label{distinguishing-capability-from-specification}

Reading the failures across agents rather than per agent proved the most
useful analysis available, because the failures separate into two kinds.
When agents fail a task in different ways --- one breaching a voltage
limit where the others hold it, one returning a wrong ride-through
verdict, one inverting a sign --- the disagreement is signal, and it is
what a leaderboard should reward. But when every agent fails a task in
the same way, the task itself, rather than the agents, is the first
thing to suspect. Treating that pattern as a diagnostic during
construction surfaced a series of specification defects --- mismatches
between a field's description and the value the evaluator expected,
unstated conventions, and over-strict comparisons that rejected
admissible answers --- each confirmed by correcting the specification
and watching the agents recover. The pattern is a suspicion, not a
verdict: a unanimous failure is sometimes a shared blind spot rather
than a defect, so it should be confirmed by inspection. We propose this
rule --- treat a unanimous failure as a likely defect in the
specification or the evaluator, then confirm or dismiss it --- as a
routine quality-control step for executable benchmarks.

The most consequential instance pointed not at a task but at an
evaluator, and is worth stating in full, as it is the clearest evidence
we have for the value of the method. On one held-out reliability case
concerning faulted-section localization, all three agents identified the
same faulted section, a conclusion consistent both with the
fault-current-indicator readings and with the pattern of customer
complaints; the evaluator demanded a different section that in fact
contradicted the case's own indicator readings. The agents were correct
and the evaluator was wrong. The cause was an off-by-one error in the
evaluator's indicator model: it had placed the faulted section's
downstream node on the fault-current path, so that an indicator at that
node was predicted to register a fault current it would not in fact see,
whereas an indicator registers fault current only when the fault lies
strictly downstream of it. What makes this more than a corrected bug is
that the benchmark's own safeguards could not have caught it. The
reference solver used for self-checking is derived from the same
function as the evaluator, so solver and evaluator agreed with each
other and the public sample passed; the error was invisible from within
and became visible only because three independent agents brought the
correct physics to bear. An executable benchmark, in other words, can
harbor a latent domain error in the very code that defines correctness,
and a matrix of agents is a means of finding it --- a caution we would
extend to any benchmark whose reference answers are derived from its
evaluator.

\subsubsection{8.3 Held-out generator
validation}\label{held-out-generator-validation}

The held-out mechanism of Section 4.4 is only useful if the generator
reliably produces solvable, varied instances rather than near-duplicates
of the public case. We tested this directly by drawing eight held-out
variants of every family from a fixed validation seed and scoring each
with the reference solver. All 328 variants (41 families \(\times\) 8)
were produced without exhausting the redraw budget and were feasible, so
the acceptance rule --- keep a draw only if the reference solver solves
it and no small perturbation of that solution scores higher --- holds at
scale. Every variant differed from its public case, and 40 of the 41
families yielded eight distinct instances; the exception,
faulted-section localization, draws a fault location from a small
discrete set and so produces fewer distinct instances (four of eight)
--- an inherent bound for categorical families rather than a generator
fault. Across the continuous content the perturbed inputs varied with a
mean coefficient of variation near 0.2, so the variants are genuinely
different problems. This validates the generator itself: it is a
property of the held-out construction, not a measure of agent
performance, for which repeated agent runs over many variants --- beyond
the single configuration reported above --- would be required.

\subsubsection{8.4 A wider spectrum of models and
harnesses}\label{a-wider-spectrum-of-models-and-harnesses}

The three reference agents each bundle a harness with a model --- Codex
with gpt-5.5, Cursor with composer-2.5, OpenCode with the free
deepseek-v4-flash. To probe both axes, we ran a small grid on the public
split: two model-agnostic harnesses, OpenCode and Aider, each driving
four further, widely accessible models, with every cell scored by the
same optimum-normalized evaluators. OpenCode therefore appears both in
the reference round (on the free model) and here (on the four others).

\begin{longtable}[]{@{}lll@{}}
\caption{Public-split mean score for two model-agnostic harnesses across
four models.}\tabularnewline
\toprule\noalign{}
model & OpenCode & Aider \\
\midrule\noalign{}
\endfirsthead
\toprule\noalign{}
model & OpenCode & Aider \\
\midrule\noalign{}
\endhead
\bottomrule\noalign{}
\endlastfoot
Kimi K2.5 (\textasciitilde1T MoE, 32B active, open) & 0.93 & 0.85 \\
GLM-4.6 (357B, 32B active, open) & 0.78 & 0.44 \\
Claude Haiku 4.5 (proprietary) & 0.71 & 0.51 \\
Qwen3-Coder-Next (80B, 3B active, open) & 0.66 & 0.53 \\
\end{longtable}

Two things stand out. Under OpenCode the public means span 0.66 to 0.93,
well below the reference frontier at 0.95--0.97, so the benchmark
separates the realistic spectrum of models a practitioner might deploy,
and the ordering tracks more than raw size --- the two 32B-active models
differ by 0.15. And the harness matters as much as the model: every
model scores lower under Aider, a code-editing harness applied here to
an answer-writing task, by as much as 0.34 for GLM-4.6. The same model
is thus a different agent under different scaffolding --- the
agent-agnostic contract put to use, and the reason we keep the harness,
not only the model, in view (Section 4.2). Caveats keep this honest:
each cell is a single run; the lower Aider scores are overwhelmingly
genuine wrong answers, not format errors (one empty submission aside);
two OpenCode runs had a few infrastructure no-submission misses; and of
the other harnesses we tried, Forge and Terminus were unavailable in our
environment and Goose would not route to our model gateway. The genuine
failures are of the kind the benchmark is meant to catch: a fault
current reported a thousand times too small, dispatch schedules that
breach state-of-charge limits, a missed islanding flag, an inverted
ride-through verdict.

\subsection{9. Limitations}\label{limitations}

Several limitations bound what these results support. The public cases
are compact --- networks of a few buses and short time series --- so
they test whether an agent commands a method, not whether it scales to
industrial data volumes. The evaluators are deterministic surrogates,
and no claim of industrial AC accuracy is warranted until
simulator-backed checkers are in place. With one case per family per
split, a ranking over 41 points is statistically fragile, though the
generator makes enlarging the sample inexpensive. The tasks are also
self-contained: an agent computes each answer directly and is never
required to operate professional power-system software, so the benchmark
does not yet measure mastery of the simulators and tools that much real
engineering work depends on. A reference solution scores one by
construction, which is a self-consistency check on the contract, not a
lower bound on difficulty, and cannot validate the evaluator. The
quality-control method of Section 8.2 --- reading a unanimous agent
failure as a likely defect --- has a limit that grows with the
benchmark's difficulty: it can tell a broken task from a merely hard one
only while the agents are strong enough that genuine failures are rare,
so as tasks approach the frontier and honest failures become common, a
unanimous failure no longer points clearly at the specification, and the
unsolved tail must then be validated by independent recomputation rather
than by agent agreement. And because each score is reported relative to
its task's reference optimum, the scale is only as trustworthy as that
reference's optimality: we verify optimality when a held-out case is
generated and treat any solution that beats the reference as a flag
rather than a score above one, but we do not prove global optimality for
every family, so any future continuous-objective task would need its
reference established with the same care. The seed of a held-out round
is spent once the round is disclosed, so a citable leaderboard requires
a fresh seed on an isolated host. The sandbox isolates by relocation
rather than containerization, which removes leakage through relative
paths and environment but not a determined search for the evaluator by
absolute path, so competitive use would want host-level containers;
packaging the benchmark as an environment for a container-based
evaluation harness such as Harbor \citep{harbor2026} is a natural route
to that isolation, and would additionally supply the parallel execution
and rollout interfaces needed to estimate the sampling variance we
currently leave unmeasured. We run one configuration per agent and so do
not estimate the variance of model sampling.

A broader question is what the benchmark is worth when the strongest
current agents already solve most of its public tasks. On this compact
tier its power to rank the leading agents is limited, and we do not
present it as a frontier-stumping leaderboard; its value rests
elsewhere. It separates the realistic spectrum of agents rather than
only the leaders, with OpenCode, on a small open model, trailing the
frontier clearly and failing outright where the leaders succeed. This
realistic tier is where the benchmark is most useful in practice: cost,
data-governance, latency, offline-operation, and local-infrastructure
constraints often push practitioners toward smaller open models,
free-tier services, or corporate model gateways rather than the
strongest proprietary agents, and on exactly that tier the benchmark
retains resolution --- in the public model--harness grid, mean scores
range from 0.66 to 0.93 under OpenCode and from 0.44 to 0.85 under
Aider, with failures that are genuine engineering errors rather than
formatting artifacts. The compact tier thus acts less as a frontier
leaderboard than as a deployment filter: it can tell which non-frontier
agents are sound enough to merit simulator-backed testing and which fail
basic physically checked tasks first. It surfaces shared blind spots
even when scores are high, because a task that every agent misses in the
same way is a capability gap an executable check makes visible and a
prose grade would not. Its contract is built to grow into the harder
tiers of Section 10, where an agent must operate professional simulators
rather than compute compact instances and where the headroom is large.
And its resistance to contamination is prospective, the generator
keeping an uncontaminated draw available as the public set ages and is
trained on. The contribution is a growing, physics-grounded,
contamination-resistant yardstick for a domain that had only textual
assessment, not a one-round wall of difficulty.

\subsection{10. Conclusion}\label{conclusion}

We have brought executable evaluation, by now standard for software and
tool-use agents, to electric power engineering, where assessment had
remained textual. The benchmark turns power-system tasks into checks
that recompute physics; on a matrix of current agents the strongest
already score close to the maximum on the present compact tier while
OpenCode, a smaller open model, trails, and public and held-out
performance are consistent. Its most transferable contribution is not
any single task but a method: reading a unanimous failure as a defect
signal, which over the course of this work repeatedly uncovered
specification defects and even a latent error in an evaluator's own
physics that the benchmark could not have caught by self-consistency
alone --- a reminder that a reference answer derived from an evaluator
cannot certify it. The lasting value of the artifact lies in the
contract, the domain-grounded task set, the evaluators, the seed-driven
held-out generator with its prospective resistance to contamination, the
isolated runner, and the reproducible quality-control procedure; one
line of work ahead is to fill the evaluators with simulator-backed
implementations without disturbing the contract that holds them
together.

A second axis of growth concerns the agent rather than the evaluator.
The tasks presented here are self-contained: an agent reads a case and
computes an answer, and while it may write and run scratch code, no task
yet requires it to operate professional power-system software. Yet much
of real engineering work is exactly that --- driving a load-flow
package, a short-circuit tool, or a dynamic simulator --- and the
production agents now emerging in the domain already orchestrate such
tools. The contract imposes no limit here: an agent is free to invoke
whatever tools its sandbox provides to produce a solution. A natural
next generation of tasks would therefore measure not closed-form
competence but mastery of these specialized environments, with the
simulator serving as the agent's instrument rather than the evaluator's.
The duality is worth naming: the same simulator can sit on either side
of the contract --- the agent's tool for solving, or the evaluator's
backend for checking --- and the isolation of Section 4.2 is exactly
what must keep the two apart, so that an agent can never reach the
checking instance as an oracle.

\bibliography{references}

\footnotesize

\subsection{Appendix A. Task Catalog}\label{appendix-a.-task-catalog}

The 41 families are listed below by domain area, each with its primary
source or governing standard and the confidence level of Section 5.1 (11
high, 28 medium, 2 low). The benchmark ships the full record for every
family in English and Russian: the complete task contract --- inputs,
objective, hard constraints, and submission schema --- together with the
provenance entry (source title, locator, and the note mapping the source
onto the executable case).

\textbf{1. Network analysis and operating constraints}

\begin{longtable}[]{@{}
  >{\raggedright\arraybackslash}p{(\columnwidth - 4\tabcolsep) * \real{0.3333}}
  >{\raggedright\arraybackslash}p{(\columnwidth - 4\tabcolsep) * \real{0.3333}}
  >{\raggedright\arraybackslash}p{(\columnwidth - 4\tabcolsep) * \real{0.3333}}@{}}
\toprule\noalign{}
\begin{minipage}[b]{\linewidth}\raggedright
Family
\end{minipage} & \begin{minipage}[b]{\linewidth}\raggedright
Source / standard
\end{minipage} & \begin{minipage}[b]{\linewidth}\raggedright
Confidence
\end{minipage} \\
\midrule\noalign{}
\endhead
\bottomrule\noalign{}
\endlastfoot
\texttt{ybus\_construction} & Glover, Overbye \& Sarma (textbook) &
high \\
\texttt{pf\_voltage\_repair} & Glover, Overbye \& Sarma (textbook) &
high \\
\texttt{contingency\_screening} & Graph-computing contingency screening
& medium \\
\texttt{corrective\_transmission\_switching} & SCUC with corrective
switching (arXiv:2001.00597) & medium \\
\texttt{dynamic\_thermal\_rating} & IEEE 738; Safari et al.,
probabilistic DTR & medium \\
\end{longtable}

\textbf{2. Short circuit and protection}

\begin{longtable}[]{@{}
  >{\raggedright\arraybackslash}p{(\columnwidth - 4\tabcolsep) * \real{0.3333}}
  >{\raggedright\arraybackslash}p{(\columnwidth - 4\tabcolsep) * \real{0.3333}}
  >{\raggedright\arraybackslash}p{(\columnwidth - 4\tabcolsep) * \real{0.3333}}@{}}
\toprule\noalign{}
\begin{minipage}[b]{\linewidth}\raggedright
Family
\end{minipage} & \begin{minipage}[b]{\linewidth}\raggedright
Source / standard
\end{minipage} & \begin{minipage}[b]{\linewidth}\raggedright
Confidence
\end{minipage} \\
\midrule\noalign{}
\endhead
\bottomrule\noalign{}
\endlastfoot
\texttt{three\_phase\_short\_circuit} & IEC 60909 & high \\
\texttt{earth\_fault\_calculation} & Anderson, Analysis of Faulted Power
Systems & medium \\
\texttt{breaker\_relay\_short\_circuit} & IEC 60909; Glover, Overbye \&
Sarma & high \\
\texttt{distance\_protection\_settings} & Horowitz \& Phadke, Power
System Relaying & medium \\
\texttt{overcurrent\_protection} & Horowitz \& Phadke, Power System
Relaying & medium \\
\texttt{cable\_ampacity\_sizing} & IEC 60364-5-52 / IEC 60949 & high \\
\end{longtable}

\textbf{3. Stability, grid code, and inverter-based resources}

\begin{longtable}[]{@{}
  >{\raggedright\arraybackslash}p{(\columnwidth - 4\tabcolsep) * \real{0.3333}}
  >{\raggedright\arraybackslash}p{(\columnwidth - 4\tabcolsep) * \real{0.3333}}
  >{\raggedright\arraybackslash}p{(\columnwidth - 4\tabcolsep) * \real{0.3333}}@{}}
\toprule\noalign{}
\begin{minipage}[b]{\linewidth}\raggedright
Family
\end{minipage} & \begin{minipage}[b]{\linewidth}\raggedright
Source / standard
\end{minipage} & \begin{minipage}[b]{\linewidth}\raggedright
Confidence
\end{minipage} \\
\midrule\noalign{}
\endhead
\bottomrule\noalign{}
\endlastfoot
\texttt{critical\_clearing\_time} & Equal-area criterion (Kundur) &
high \\
\texttt{transient\_stability\_prediction} & Chen et al.,
transient-stability prediction & medium \\
\texttt{frt\_compliance} & IEEE 2800 / ENTSO-E RfG (FRT) & medium \\
\texttt{ibr\_short\_circuit\_frt} & IBR modeling for short circuit / FRT
& medium \\
\texttt{min\_synchronous\_share} & Wen et al., low-inertia frequency
stability & low \\
\end{longtable}

\textbf{4. Distributed resources, PV, EV, and storage}

\begin{longtable}[]{@{}
  >{\raggedright\arraybackslash}p{(\columnwidth - 4\tabcolsep) * \real{0.3333}}
  >{\raggedright\arraybackslash}p{(\columnwidth - 4\tabcolsep) * \real{0.3333}}
  >{\raggedright\arraybackslash}p{(\columnwidth - 4\tabcolsep) * \real{0.3333}}@{}}
\toprule\noalign{}
\begin{minipage}[b]{\linewidth}\raggedright
Family
\end{minipage} & \begin{minipage}[b]{\linewidth}\raggedright
Source / standard
\end{minipage} & \begin{minipage}[b]{\linewidth}\raggedright
Confidence
\end{minipage} \\
\midrule\noalign{}
\endhead
\bottomrule\noalign{}
\endlastfoot
\texttt{pv\_volt\_var} & Turitsyn et al., local Volt-VAR control &
medium \\
\texttt{ev\_v2g\_outage\_schedule} & EVs for power quality \& security &
medium \\
\texttt{ev\_v2g\_voltage\_support} & EVs for power quality \& security &
medium \\
\texttt{bess\_ancillary\_response} & Gonzalez-Longatt \& Rueda Torres &
medium \\
\texttt{commercial\_pv\_lcoe\_uncertainty} & PV soft-cost uncertainty
(NREL) & medium \\
\end{longtable}

\textbf{5. Microgrids and dispatch}

\begin{longtable}[]{@{}
  >{\raggedright\arraybackslash}p{(\columnwidth - 4\tabcolsep) * \real{0.3333}}
  >{\raggedright\arraybackslash}p{(\columnwidth - 4\tabcolsep) * \real{0.3333}}
  >{\raggedright\arraybackslash}p{(\columnwidth - 4\tabcolsep) * \real{0.3333}}@{}}
\toprule\noalign{}
\begin{minipage}[b]{\linewidth}\raggedright
Family
\end{minipage} & \begin{minipage}[b]{\linewidth}\raggedright
Source / standard
\end{minipage} & \begin{minipage}[b]{\linewidth}\raggedright
Confidence
\end{minipage} \\
\midrule\noalign{}
\endhead
\bottomrule\noalign{}
\endlastfoot
\texttt{microgrid\_economic\_dispatch} & España et al., microgrid
dispatch & medium \\
\texttt{rolling\_microgrid\_dispatch} & España et al., microgrid
dispatch & medium \\
\texttt{islanded\_microgrid\_pq\_dispatch} & España et al., microgrid
dispatch & medium \\
\texttt{dispatch\_uncertainty} & Chung, advanced prediction for smart
grids & low \\
\texttt{hydro\_thermal\_storage\_uc} & ADMM unit commitment & medium \\
\texttt{n1\_generation\_commitment} & SCUC with corrective switching
(arXiv:2001.00597) & medium \\
\end{longtable}

\textbf{6. Reliability and restoration}

\begin{longtable}[]{@{}
  >{\raggedright\arraybackslash}p{(\columnwidth - 4\tabcolsep) * \real{0.3333}}
  >{\raggedright\arraybackslash}p{(\columnwidth - 4\tabcolsep) * \real{0.3333}}
  >{\raggedright\arraybackslash}p{(\columnwidth - 4\tabcolsep) * \real{0.3333}}@{}}
\toprule\noalign{}
\begin{minipage}[b]{\linewidth}\raggedright
Family
\end{minipage} & \begin{minipage}[b]{\linewidth}\raggedright
Source / standard
\end{minipage} & \begin{minipage}[b]{\linewidth}\raggedright
Confidence
\end{minipage} \\
\midrule\noalign{}
\endhead
\bottomrule\noalign{}
\endlastfoot
\texttt{flisr\_restoration} & Two-stage distribution service restoration
& medium \\
\texttt{fault\_section\_localization} & Brown, faulted-circuit
indicators & medium \\
\texttt{fci\_placement} & Brown, faulted-circuit indicators & medium \\
\texttt{fci\_saidi\_caidi} & Brown, FCIs; IEEE 1366 indices & medium \\
\texttt{operator\_breaker\_load\_actions} & Glover, Overbye \& Sarma
(textbook) & medium \\
\end{longtable}

\textbf{7. Power quality, standards, assets, and cybersecurity}

\begin{longtable}[]{@{}
  >{\raggedright\arraybackslash}p{(\columnwidth - 4\tabcolsep) * \real{0.3333}}
  >{\raggedright\arraybackslash}p{(\columnwidth - 4\tabcolsep) * \real{0.3333}}
  >{\raggedright\arraybackslash}p{(\columnwidth - 4\tabcolsep) * \real{0.3333}}@{}}
\toprule\noalign{}
\begin{minipage}[b]{\linewidth}\raggedright
Family
\end{minipage} & \begin{minipage}[b]{\linewidth}\raggedright
Source / standard
\end{minipage} & \begin{minipage}[b]{\linewidth}\raggedright
Confidence
\end{minipage} \\
\midrule\noalign{}
\endhead
\bottomrule\noalign{}
\endlastfoot
\texttt{en50160\_voltage\_compliance} & EN 50160 & high \\
\texttt{power\_quality\_event\_classification} & IEC 61000-4-30 &
high \\
\texttt{harmonic\_ieee519\_compliance} & IEEE 519 & high \\
\texttt{transformer\_thermal\_loading} & IEC 60076-7 & high \\
\texttt{fdi\_state\_estimation} & Abur \& Exposito, state estimation &
high \\
\texttt{protected\_meter\_placement} & Graphical defense vs false-data
injection & medium \\
\texttt{mas\_frequency\_handover} & Multi-agent control of active
distribution networks & medium \\
\end{longtable}

\textbf{8. Forecasting under uncertainty}

\begin{longtable}[]{@{}
  >{\raggedright\arraybackslash}p{(\columnwidth - 4\tabcolsep) * \real{0.3333}}
  >{\raggedright\arraybackslash}p{(\columnwidth - 4\tabcolsep) * \real{0.3333}}
  >{\raggedright\arraybackslash}p{(\columnwidth - 4\tabcolsep) * \real{0.3333}}@{}}
\toprule\noalign{}
\begin{minipage}[b]{\linewidth}\raggedright
Family
\end{minipage} & \begin{minipage}[b]{\linewidth}\raggedright
Source / standard
\end{minipage} & \begin{minipage}[b]{\linewidth}\raggedright
Confidence
\end{minipage} \\
\midrule\noalign{}
\endhead
\bottomrule\noalign{}
\endlastfoot
\texttt{wind\_power\_forecast} & Safari et al., short-term wind
forecasting & medium \\
\texttt{wind\_prediction\_interval} & Khorramdel et al., fuzzy
prediction intervals & medium \\
\end{longtable}

\subsection{Appendix B. Experiment
Artifacts}\label{appendix-b.-experiment-artifacts}

\textbf{Run configuration.} Each agent was invoked once per case through
its own command-line interface under a 600-second per-case timeout and
the infrastructure-retry policy of Section 4.2 --- at most two attempts,
retrying only provider- or transport-transient failures, never quality,
contract, or timeout failures. One public case and one freshly generated
held-out case were run per family, and every submission was scored by
the deterministic, optimum-normalized evaluators of Section 6. The
versions were Codex on gpt-5.5 at medium reasoning effort (codex-cli
0.139.0), Cursor's command-line agent on composer-2.5 (cursor-agent
2026.06.12-01-15-52-7244546), and OpenCode on deepseek-v4-flash-free
(opencode 1.15.13).

\textbf{Public shortfalls.} Every public case not scored one is as
follows. Codex missed contingency\_screening (0.9) and returned an
infeasible ev\_v2g\_outage\_schedule, with pv\_volt\_var within
\(10^{-5}\) of the optimum but short of an exact one; Cursor missed
contingency\_screening (0.9) and ev\_v2g\_outage\_schedule (infeasible);
OpenCode returned infeasible solutions for ev\_v2g\_outage\_schedule and
pv\_volt\_var. All remaining public cases scored one.

\textbf{Distribution.} The reference round ships with the benchmark as
per-agent, per-split run summaries. For the public split the per-case
submissions, evaluator results, and agent logs are included; for the
held-out split only the aggregate summaries are published, since a
per-case held-out submission or evaluator result would disclose that
case's answer (Section 4.4). The wider model--harness grid of Section
8.4 ships likewise, as per-cell public submissions with their recomputed
per-case results and summaries.

\end{document}